\documentclass[sigconf]{acmart}
\usepackage{fancyhdr}
\AtBeginDocument{%
  }

\setcopyright{acmlicensed}
\usepackage{multirow}
\usepackage{adjustbox}
\usepackage{fixltx2e}
\usepackage{graphicx}
\usepackage{multicol}
\copyrightyear{2018}
\acmYear{2018}
\acmDOI{XXXXXXX.XXXXXXX}
\acmISBN{978-1-4503-XXXX-X/2018/06}
\renewcommand\footnotetextcopyrightpermission[1]{}
\makeatletter

\newcommand{\Rmnum}[1]{\expandafter\@slowromancap\romannumeral #1@}
\makeatother

\begin{document}
\thispagestyle{fancy}  
\begin{sloppypar}
\title{ConFusion: Continuous Fusion Space Learning for Fine-Grained Controllable Infrared and Visible Image Fusion}

\author{Yurong Guo}
\affiliation{%
  \institution{North China Electric Power University}
  \city{Baoding}
  \state{Hebei}
  \country{China}
}
\email{guoyurong@ncepu.edu.cn}

\author{Yufei He}
\affiliation{%
  \institution{North China Electric Power University}
  \city{Baoding}
  \state{Hebei}
  \country{China}
}
\email{220232215006@ncepu.edu.cn}

\author{Yonghao Li}
\affiliation{%
  \institution{North China Electric Power University}
  \city{Baoding}
  \state{Hebei}
  \country{China}
}
\email{220252215115@ncepu.edu.cn}

\author{Dongliang Chang}
\affiliation{%
  \institution{Beijing University of Posts and Telecommunications}
  \city{Beijing}
  \country{China}
}
\email{changdongliang@bupt.edu.cn}

\author{Ke Zhang}
\authornote{Corresponding Author.}
\affiliation{%
  \institution{North China Electric Power University}
  \city{Baoding}
  \state{Hebei}
  \country{China}
}
\email{zhangkeit@ncepu.edu.cn}

\author{Zhanyu Ma}
\affiliation{%
  \institution{Beijing University of Posts and Telecommunications}
  \city{Beijing}
  \country{China}
}
\email{mazhanyu@bupt.edu.cn}


\begin{abstract}
    Controllable infrared–visible image fusion aims to integrate complementary thermal and structural information with flexible region-aware modulation, producing fused images that adapt to diverse user requirements and downstream tasks.
    However, existing methods typically rely on predefined discrete control conditions, leading to a sparse space that fails to support fine-grained modulation demands.
    To address this, we propose ConFusion, a novel framework that learns the continuous fusion space via Gaussian-conditioned spatial-aware modulation, enabling instance-level fine-grained controllable infrared and visible image fusion.
    ConFusion employs a dual-branch architecture to disentangle modality-invariant and modality-specific representations under joint reconstruction and text-guided semantic alignment. 
   Gaussian-conditioned instance modulation variables coupled with Grounded SAM-based instance masks guide instance-level fine-grained modulation through the Mask-Guided Specific Feature Modulator, while the Text-Driven Invariant Feature Enhancer improves semantic consistency and enhances fusion. 
    During inference, the multimodal large language model parses user intents into instance-level modulation variables to guide image fusion. 
    Extensive experiments show that ConFusion achieves state-of-the-art performance across multiple metrics in both fusion quality and downstream tasks, while supporting fine-grained controllable image fusion. Our code is available at https://github.com/HeyufeiAnto/Confusion

\end{abstract}


\maketitle

\section{Introduction}
Image fusion integrates complementary information from multiple modalities to generate a fused image with enriched details~\cite{review2}.
Among various paradigms, infrared–visible image fusion (IVIF) has attracted considerable attention due to its broad applicability in real-world scenarios~\cite{iffdcta}. 
Infrared images capture thermal radiation and remain robust under low-light or nighttime conditions but typically lack fine-grained texture, whereas visible images provide rich structural details yet are sensitive to illumination variations~\cite{fusiongan}. 
Therefore, IVIF aims to preserve salient thermal cues while incorporating detailed visual information, enabling more informative and reliable fused representations~\cite{review}.
It has been widely applied in downstream applications, including assisted driving~\cite{dcevo}, remote sensing~\cite{remote}, surveillance~\cite{mif}, and target detection~\cite{detfusion}.

\begin{figure}
\includegraphics[width=0.9\linewidth]{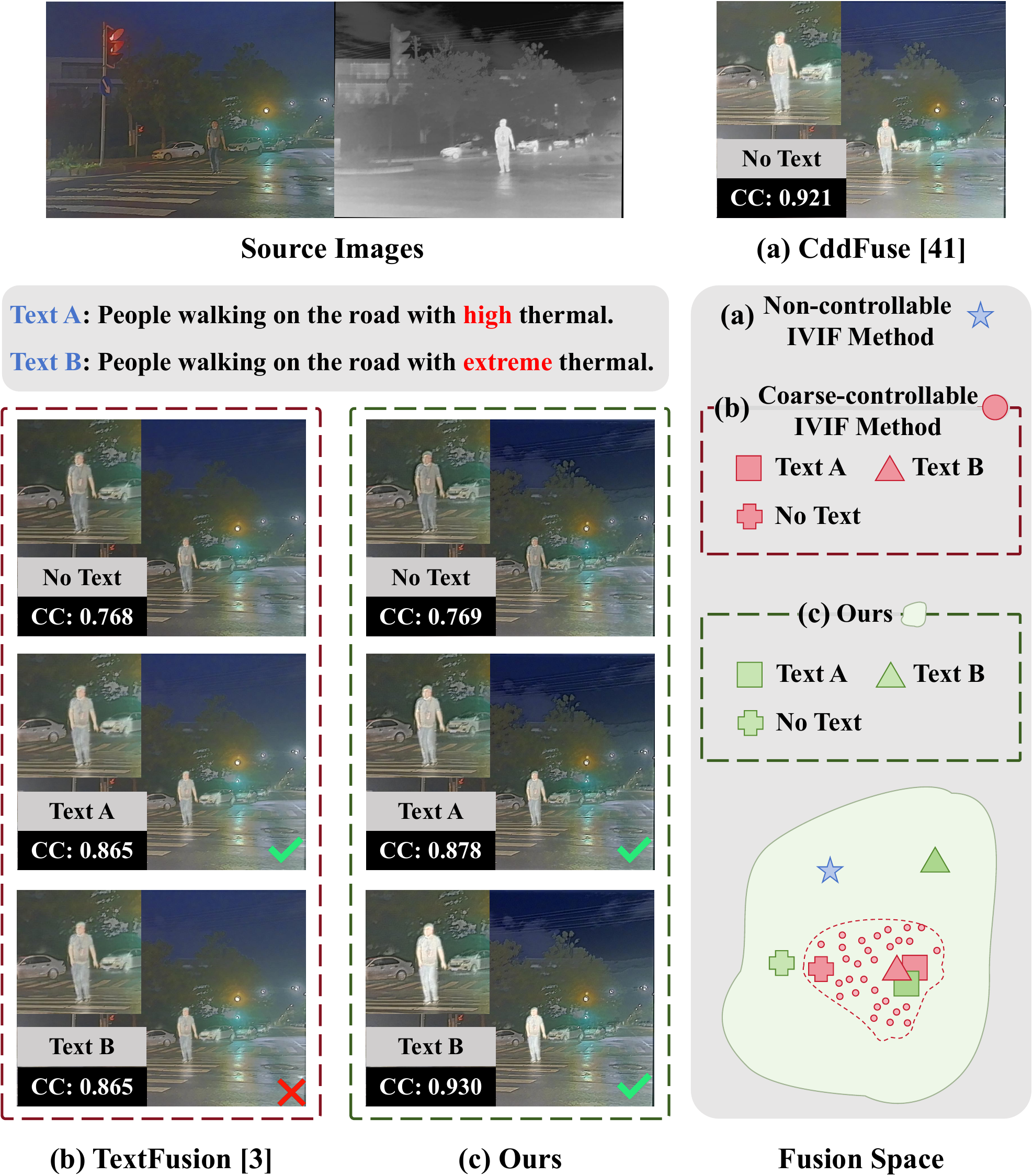}
\caption{(a) Non-controllable IVIF: learns a deterministic one-to-one mapping, producing a single fixed fusion output.
(b) Coarse-controllable IVIF: learns a sparsely sampled fusion space, yielding consistent instance-level responses under subtly different text inputs.
(c) Ours: learns a continuous fusion space, producing instance-level fusion results that vary with fine-grained text inputs.
CC denotes the correlation coefficient between fused and infrared images in instance regions, where higher values indicate stronger thermal representation.
}
\vspace{-0.2cm}
\label{fig1}
\end{figure}
\label{sec:intro}
Recent advances in deep learning have substantially improved IVIF performance.
Conventional IVIF methods primarily focus on generating high-quality fused images to improve visual clarity and detail for human perception~\cite{densefuse, nestfuse}, as well as enhance downstream task performance~\cite{tardal, segmif}. 
Due to the absence of physical ground-truth fused images, these approaches rely on carefully designed network architectures optimized with mathematically defined loss functions~\cite{stdfusionet} to constrain the fusion process. 
Despite achieving impressive visual results, they generally learn a deterministic one-to-one mapping that produces a single fixed fusion output for each input pair, as illustrated in Figure~\ref{fig1}(a). 
Consequently, such a paradigm leads to inherently \textit{non-controllable fusion}, failing to accommodate diverse user preferences and task-specific requirements.

Multimodal large language models (LLMs) have recently emerged as a promising paradigm for incorporating user intents into controllable fusion.
Existing approaches, which we refer to as \textit{coarse-controllable IVIF fusion}, encode user requirements as predefined text or visual prompts and map them into a discrete set of control representations, enabling instance-level controllable fusion.
However, such discrete conditioning leads to a sparse sampled controllable fusion space, limiting its ability to adapt to fine-grained user intents.
As illustrated in Figure~\ref{fig1}(b), TextFusion, a representative coarse-controllable fusion method, enhances infrared features in semantically specified regions (e.g., “people”) via language instructions.
However, it produces consistent instance-level responses under subtly different intents (Text A $vs.$ Text B), as evidenced by identical Correlation Coefficient (CC) values, due to their collapse to the same point in the fusion space.

In this paper, we propose \textit{ConFusion}, a novel framework that learns a \textit{continuous fusion space via Gaussian-conditioned spatial-aware modulation}, enabling instance-level fine-grained controllable infrared–visible image fusion.
Specifically, ConFusion adopts a dual-branch architecture to explicitly disentangle modality-invariant and modality-specific representations, supervised by joint reconstruction and text-guided invariant semantic alignment.
For modality-specific features, we integrate Gaussian-sampled instance-level variables with Grounded-SAM-based instance masks to construct fine-grained spatial-aware modulation masks, and further design a Mask-Guided Specific Feature Modulator (MGSM) to achieve precise feature modulation.
For modality-invariant features, we introduce the Text-Driven Invariant Feature Enhancer (TDIE) module to improve semantic consistency and enhance fusion.
During inference, an LLM parses user intents into instance-level modulation variables along with corresponding fine-grained spatial-aware modulation masks, which are injected into the learned ConFusion to generate intent-aligned fusion images.
As illustrated in Fig.~\ref{fig1}(c), ConFusion learns a continuous latent fusion manifold, where diverse text inputs correspond to appropriate solutions in the fusion space, generating instance-level fusion results that reflect fine-grained text-driven modulation.
In summary, the main contributions of this paper are threefold:
\begin{enumerate}
   \item we propose ConFusion, which learns a continuous fusion space via Gaussian-conditioned spatial-aware modulation, enabling instance-level fine-grained controllable infrared and visible image fusion.
    
    \item We design a dual-branch feature disentanglement with text-guided semantic alignment, with a Mask-Guided Specific Feature Modulator for modality-specific features and a Text-Driven Invariant Feature Enhancer for modality-invariant features, achieving precise controllable fusion.

    \item Extensive experiments show that ConFusion achieves state-of-the-art performance across multiple metrics in both fusion quality and downstream tasks.

\end{enumerate}

\section{Related Works}
\subsection{Non-controllable IVIF Method}
With the advancement of deep learning~\cite{ding2021ap, li2020oslnet,chang2021your, guo2024understanding}, data-driven methods have become mainstream in image fusion~\cite{review3}. Early works mainly focused on improving visual quality. Li et al.~\cite{densefuse} introduced DenseBlock to extract multimodal features and adopted weighted fusion strategies, supervised by L$2$ and SSIM~\cite{ssim} Losses to preserve information from both modalities. Building upon this foundation, RFN-Nest~\cite{rfnnest} replaced hand-crafted fusion rules with learnable ResNet-based~\cite{resnet, wu2023bi, guo2022learning, guo2023task} modules, enabling adaptive feature fusion in an end-to-end manner. To achieve more refined fusion, Ma et al.~\cite{stdfusionet} leveraged salient object detection to separate foreground and background regions, and designed region-specific loss functions to emphasize thermal targets and background details, respectively. However, these methods apply global fusion without distinguishing different feature types. In contrast, Zhao et al.~\cite{cddfuse} proposed a dual-branch encoder with decoupled loss functions to disentangle multimodal features into shared background information and modality-specific details, improving both interpretability and fusion performance.

Despite achieving visually pleasing results, these methods often suffer from a domain gap between fusion and downstream tasks, leading to suboptimal performance in applications such as object detection and semantic segmentation~\cite{du2024demofusion}. To mitigate this issue, recent works incorporated task supervision into fusion.
TarDAL~\cite{tardal} jointly optimized generative adversarial networks~\cite{GAN} and YOLOv5, encouraging fused images to benefit detection. However, directly introducing detection loss provides limited improvement. To further reduce the gap, Sun et al.~\cite{detfusion} integrated the backbone of Faster R-CNN~\cite{Fasterrcnn} into the fusion framework and use task-driven attention maps to guide representation learning.
Beyond these methods, recent works explore feature interaction between fusion and task models to enable joint representation learning.
Liu et al.~\cite{metafusion} introduce a meta-feature embedding module to generate target-aware semantic features, enhancing the alignment between fusion and detection representations. Furthermore, Bai et al.~\cite{Lossfusion} adopt a meta-learning framework to dynamically adjust modality weights in the fusion loss based on downstream task losses.
However, these methods focus on visual quality or downstream tasks and are inherently non-controllable, failing to satisfy diverse user intents.
\vspace{-0.2cm}
\subsection{Controllable IVIF Method}
Recent advances in multimodal large language models make text-driven control over image fusion possible. Leveraging CLIP's~\cite{CLIP} image-text alignment capability, Wang et al.~\cite{LDFusion} encoded fused images and textual prompts with CLIP encoders and introduced a language-driven loss to align them, generating fused images that meet text prompt requirements.
Yi et al.~\cite{textif} further integrated text into feature learning by interacting CLIP-encoded degradation descriptions with image features, improving adaptation to diverse degradation scenarios.
Beyond global control, recent methods explore instance-level guidance. 
OmniFuse~\cite{ditfuse} utilized Grounded SAM to localize text-described instances and enhances them via visual regularization and semantic reverse guidance.
Wang et al.~\cite{risfuse} designed Manifold-Based Feature Alignment Module for image–text alignment and Instance Selection Module to enable single-instance enhancement across viewpoints.
Li et al.~\cite{omnifuse} leverage Diffusion-Transformer (DiT)~\cite{dit} for enhanced text-guided fusion, jointly optimizing it on diverse datasets and tasks to generate both task-adaptive fused images and task-specific outputs.

\begin{figure*}
\includegraphics[width=0.9\linewidth]{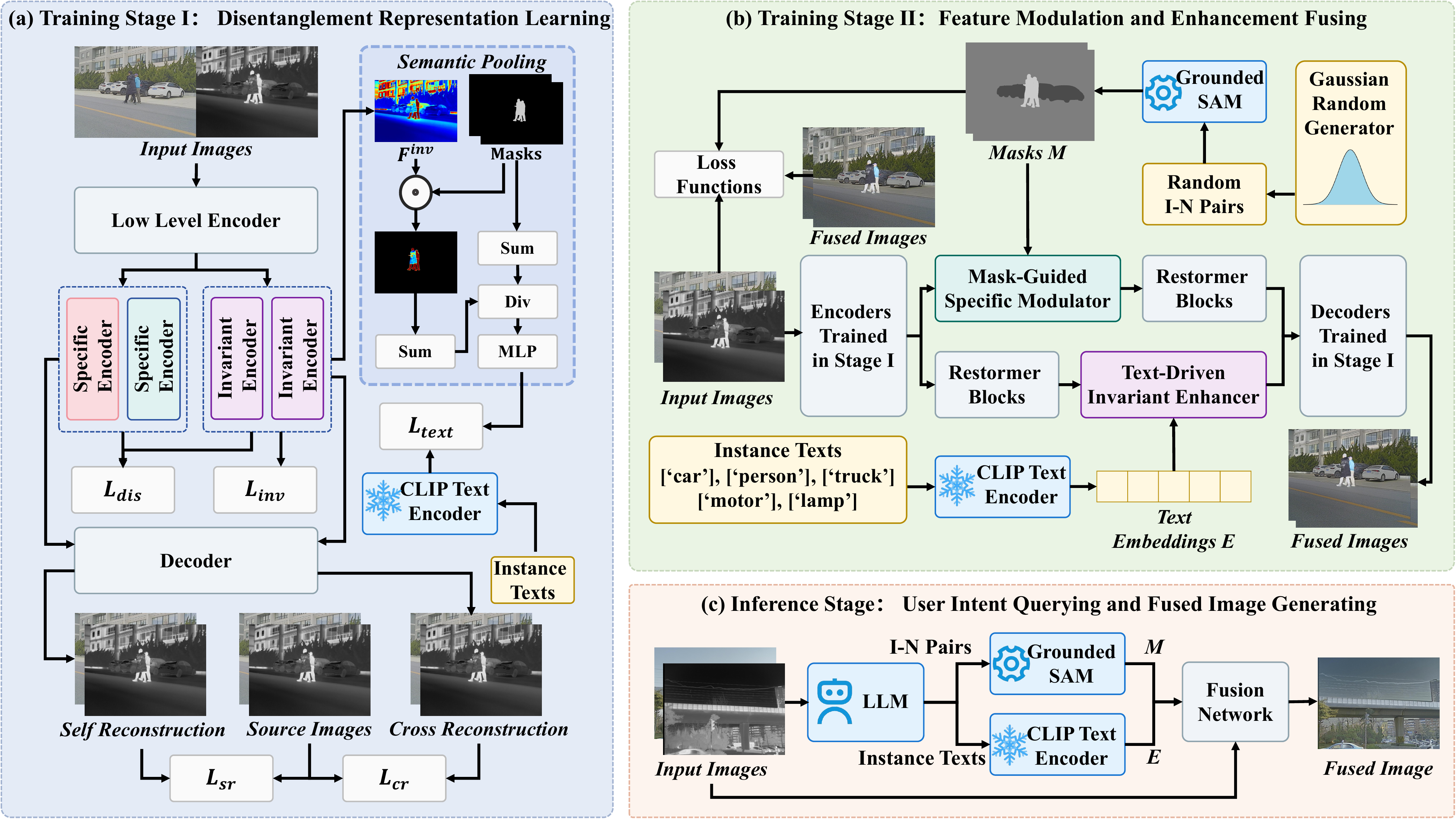}
\caption{Overall framework of the proposed ConFusion for instance-level fine-grained controllable IVIF.}
\label{fig2}
\end{figure*}

These methods improve adaptability to user intent and extend the IVIF paradigm. However, they are inherently limited to coarse controllability, as they rely on discrete conditioning based on predefined text or visual prompts, leading to a sparse controllable fusion space that struggles to satisfy fine-grained user demands.
In this paper, we focus on learning a continuous fusion space for fine-grained controllable infrared and visible image fusion.


\section{Methodology}
\subsection{Problem Formulation}
\label{sec31}
Let $I_{ir}, I_{vi} \in \mathbb{R}^{H \times W \times C}$ denote the infrared and visible images. 
Traditional fusion algorithms learn the predefined fusion function $\mathcal{F}_{if}$ to produce an immobilized fused image: $I_f = \mathcal{F}_{if}(I_{vi}, I_{ir})$. 
To improve user adaptability, recent works extend this paradigm by learning an instance-level controllable fusion function $\mathcal{F}_{if}^{\prime}$: $I_f = \mathcal{F}_{if}^{\prime}(I_{vi}, I_{ir}, p|\Omega)$, where $p\in\Omega$. $\Omega$ denotes a finite set of control embeddings introduced during training, typically constructed from predefined text or image prompts. 
Although $\Omega$ provides multiple outputs conditioned on $p$, it still remains a coarse controllable fusion.

In this paper, we propose the continuous fusion space learning method (ConFusion) for fine-grained controllable infrared and visible image fusion.
The instance-level controllable fusion function $\mathcal{F}_{if}^{\prime\prime}$ is formulated as:
\begin{equation}
I_f = \mathcal{F}_{if}^{\prime\prime}(I_{vi}, I_{ir}, p|D),
\label{eq1}
\end{equation}
where $p \sim D $ denotes instance-level modulation variables independently sampled from a Gaussian distribution. 
During training, $p$ is mapped to spatial masks and injected into the latent space to modulate modality-specific representations. Training with densely sampled instance-level modulation variables allows the model to learn a continuous and dense fusion space, supporting instance-level fine-grained controllable fusion.

\subsection{Disentanglement Representation Learning}
In this section, we introduce Training Stage \Rmnum{1} of our method. As shown in Figure~\ref{fig2}(a), we employ Feature Disentanglement and Reconstruction to effectively decouple modality-invariant and modality-specific features. Meanwhile, we leverage Invariant Semantic Alignment to strengthen the modality-agnostic of the invariant features, thereby reinforcing the disentanglement process.

\subsubsection{Feature Disentanglement and Reconstruction} 
As illustrated in Figure~\ref{fig2}(a), we first employ a shared low-level encoder to extract shallow features, followed by a dual-branch architecture to disentangle modality-specific and modality-invariant representations. 
Specifically, two modality-specific encoders learn domain-specific features $F_{ir}^{sp}$ and $F_{vi}^{sp}$ such as thermal radiation and fine-grained textures, while a shared encoder produces modality-invariant features $F_{ir}^{in}$ and $F_{vi}^{in}$ in infrared and visible images.
A decoder $\mathcal{D}$ is then introduced to reconstruct images from the disentangled features via both self-reconstruction and cross-reconstruction:
\begin{equation}
    I^{sr}_{m} = \mathcal{D}\left({\textit{concat}}\left(F^{in}_{m}, F^{sp}_{m}\right)\right), \quad I^{cr}_{m} = \mathcal{D}\left(\textit{concat}\left(F^{in}_{\bar{m}}, F^{sp}_{m}\right)\right),
\end{equation}
where $m, \bar{m}\in\{ir, vi\}$ and $m\neq\bar{m}$.
$I_{ir}^{sr}$ and $I_{ir}^{cr}$ represent the self-reconstructed and cross-reconstructed infrared images, respectively.
$I_{vi}^{sr}$ and $I_{vi}^{cr}$ are defined in the same manner for visible images.
$\textit{concat}$ denotes channel-wise concatenation.

To optimize the disentangled representations, we encourage statistical independence between invariant and specific features via a disentanglement loss:
\begin{equation}
    L_{dis} = \frac{1}{r^2} \|V_{vi}\|_F^2 + \frac{1}{r^2} \|V_{ir}\|_F^2, \quad \quad V_m =   F^{in}_m (F^{sp}_m)^T,
\end{equation}
where $m \in \{ir, vi\}$. $V_m$ denotes the cross-covariance between invariant and specific features. 
$\| \cdot \|_F$ denotes the Frobenius norm. Minimizing the Frobenius norm suppresses their linear correlations, encouraging the invariant features to be independent of modality-specific variations. In experiments, $V_m$ is approximated via random spatial sampling with $r$ sampled locations from $F^{in}_m$ and $F^{sp}_m$ for efficiency.
We further impose the consistency constraint on invariant features across modalities:
\begin{equation}
    L_{inv} = {MSE}(F_{ir}^{in}, F_{vi}^{in}),
\end{equation}
which encourages the capture of shared scene semantics. ${MSE}(\cdot)$ denotes the mean square error operator.
To ensure faithful reconstruction, we adopt a self-reconstruction loss:
\begin{equation}
    L_{sr} = \sum_{m \in \{ir, vi\}} {MSE}(I_m^{sr}, I_m) + \lambda \|\nabla I_{vi}^{sr} - \nabla I_{vi}\|_1,
\end{equation}
where $\|\cdot\|_1$ is the $L_1$ norm, and $\lambda$ is the hyperparameter. $\nabla$ denotes the spatial gradient operator. Minimizing the distance between self-reconstructed images and the source image encourages the reconstructed image to preserve the detailed features from the source image.
In addition, we apply a cross-reconstruction loss:
\begin{equation}
    L_{cr} = \sum_{m \in \{ir, vi\}} {MSE}(I_m^{cr}, I_m),
\end{equation}
which enforces consistency under invariant feature swapping.

\subsubsection{Invariant Semantic Alignment}
To further facilitate feature disentanglement, we regularize the invariant features with modality-agnostic semantic priors. Specifically, instance texts are encoded to embeddings $E$ by the CLIP encoder, which serve as semantic anchors to guide the invariant feature $F^{in}_{m}$ in capturing intrinsic instance semantics while remaining invariant to modality-specific variations. $E=\{E_j\}_{j=1}^K$, $F^{in}_{m} \in \{F_{ir}^{in}, F_{ir}^{vi} \}$ and $E\in\mathbb{R}^{K \times d}$. $K$ denotes the total number of instances present in the input image.

As illustrated in Figure~\ref{fig2}(a), we introduce a semantic pooling strategy. Given spatial masks $R_j$ generated by Grounded-SAM for $j$-th instances, we perform masked average pooling over $F^{in}_{m}$ and project the pooled features into the text embedding space:
\begin{equation}
    \hat{F}^{in}_{m,j} = \text{MLP}\left( \frac{\sum_{x,y} \left( F^{in}_{m}(x, y) \odot R_j(x, y) \right)}{\sum_{x,y} R_j(x, y) + \epsilon} \right),
\end{equation}
where $j \in \{1, 2, \dots, K\}$ and $R_j \in \mathbb{R}^{1 \times H \times W}$. ${MLP}$ denotes Multi-layer Perception. $\odot$ denotes the Hadamard product and $\epsilon$ is a small constant to prevent zero division. We then align the projected visual features $\hat{F}^{in}_{m,j}$ with their corresponding text features $E_j$:
\begin{equation}
    L_{text} = \frac{1}{2K} \sum_{m \in \{ir, vi\}} \sum_{j=1}^{K} \left( 1 - \frac{\hat{F}^{in}_{m,j} \cdot E_j}{\|\hat{F}^{in}_{m,j}\|_2 \|E_j\|_2} \right),
\end{equation}
which constrains $F^{in}_m$ to align with semantic priors, thereby reducing ambiguity between invariant and specific features and facilitating more effective disentanglement.
The overall objective function for the Training Stage \Rmnum{1} is:
\begin{equation}
    L_{stage1} = L_{inv} + L_{sr} + L_{cr} + \alpha_1 L_{text} + \alpha_2L_{dis},
\end{equation} 
where $\alpha_1$ and $\alpha_2$ are hyperparameters balancing the loss terms.

\subsection{Feature Modulation and Enhancement Fusing}
In this section, we introduce Training Stage \Rmnum{2}, as shown in Figure~\ref{fig2}(b). We adopt Mask-Guided Feature Modulation to achieve fine-grained instance-level modulation of modality-specific features, while we employ Text-Driven Invariant Feature Enhancement to improve semantic consistency and enhance fusion.
\subsubsection{Mask-Guided Specific Feature Modulation}
For modality-specific features, we propose the Gaussian-conditioned spatial-
aware modulation strategy. 
As illustrated in Figure~\ref{fig2}(b), modulation intensity variable $\alpha_j$ is sampled for each instance $o_j$ in the input image from Gaussian distribution, forming instance–intensity pairs ${\text{\{ (}}{o_j},{\alpha _j})\} _{j = 1}^K$, where $\alpha_j \sim \mathcal{N}(\mu, \sigma^2)$ serves as the instance-level modulation variable. In experiments, we empirically set it to a truncated Gaussian distribution $\mathcal{N}_{[0,1]}(0.5, 0.2^2)$.
Grounded-SAM~\cite{SAM} is then employed to obtain the mask $S_j$ for each instance, based on which the spatial-aware modulation mask is constructed as:
\begin{equation}
    M(x, y) = \sum_{j=1}^{K} \alpha_j \cdot S_j(x, y) + \alpha_{bg} \cdot S_{bg}(x, y),
\end{equation}
where $S_{bg}$ represents the mask in background region, and $\alpha_{bg}$ is the default modulation intensity for the background.

Furthermore, we propose the Mask-Guided Specific Feature Modulator (MGSM), which projects the modulation mask into the embedding space via the convolution operation to modulate the modality-specific features. Specifically, for the infrared branch, the modulation is formulated as:
\begin{equation}
f_{\omega}^{ir}, f_{\beta}^{ir} = {Chunk}({Conv}(M)),
\end{equation}
\begin{equation}
\hat{\!F}_{ir}^{sp} = f_{\omega}^{ir} \odot {LN}(F_{ir}^{sp}) + f_{\beta}^{ir}.
\end{equation}
For the visible branch, the modulation is performed using the complementary mask:
\begin{equation}
f_{\omega}^{vi}, f_{\beta}^{vi} = {Chunk}({Conv}(1 - M)),
\end{equation}
\begin{equation}
\hat{\!F}_{vi}^{sp} = f_{\omega}^{vi} \odot {LN}(F_{vi}^{sp}) + f_{\beta}^{vi},
\end{equation}
where ${Chunk}(\cdot)$ denotes the feature channel chunking operation. ${Conv}(\cdot)$ represents the convolutional layer and ${LN}(\cdot)$ stands for layer normalization. 
$\hat{\!F}_{ir}^{sp}$ and $\hat{\!F}_{vi}^{sp}$ represent the modulated specific features. 
The features are then fused via Restormer blocks~\cite{restormer} ${{Restormer}}( \cdot )$ to generate modulated modality-specific fusion features:
\begin{equation}
    F_{mod}^{sp} = {{Restormer}}(\textit{concat}(\hat{\!F}_{ir}^{sp},\hat{\!F}_{vi}^{sp})).
    \label{eq:Restormer}
\end{equation}

To further enforce region-aware modulation, we introduce the ROI-constrained loss:
\begin{equation}
\begin{split}
L_{ROI} & = \frac{1}{|M_o|} ( | M_o \odot M \odot (I_{fu} - I_{ir}) |_2^2 \\
& + | M_o \odot (1 - M) \odot (I_{fu} - I_{vi}) |_2^2  \\
& + | M_o \odot (\nabla I_{fu} - \max(\nabla I_{ir}, \nabla I_{vi})) |_1 ),
\end{split}
\end{equation}
where $M_o$ denotes the instance-region binary mask.  
$I_{fu}$ is the fused image reconstructed by enhanced modality-invariant fusion feature $F{_{enh}^{in}}$ and modulated modality-specific fusion features $F{_{mod}^{sp}}$. The enhanced modality-invariant fusion feature $F{_{enh}^{in}}$ will be introduced in \ref{sec TDSSS}.
The first and second terms constrain the fused image to present pixel intensities from the infrared and visible images in different instance regions, and the third term forces the fused image to preserve the edges with the larger gradient from two modalities within the instance region.
This loss encourages the model to modulate the expression of different modality information across instance regions in the generated fused images.

\subsubsection{Text-Driven Invariant Feature Enhancement}
\label{sec TDSSS}
To enhance semantic consistency and stabilize cross-modal representations, we propose a Text-Driven Invariant Feature Enhancer (TDIE) that injects text priors into invariant features.
Building upon the semantics alignment between invariant features and text embeddings extracted by the CLIP text encoder learned in Training Stage \Rmnum{1}, we first fuse modality-invariant feature $F_{ir}^{in}$ and $F_{vi}^{in}$ via Restormer blocks to obtain the fused invariant feature $F_{fu}^{in}$ as defined in Eq.~\eqref{eq:Restormer}, which then interacts with text embeddings through cross-attention to generate enhanced modality-invariant fusion feature.
Specifically, $F_{fu}^{in}$ is flattened into querys ${F_{que}^{in}}\in{\mathbb{R}^{HW \times C}}$, while the text embeddings $E$ are projected into keys ${E_{key}}\in{\mathbb{R}^{K \times C}}$ and values ${E_{val}}\in{\mathbb{R}^{K \times C}}$. The attention map is computed as:
\begin{equation}
A_{fu} = {Softmax}\left( \frac{F_{que}^{in} (E_{key})^T}{\sqrt{C}} \right).
\end{equation}
This map precisely highlights feature points with strong correlation between invariant features and text embeddings, which captures modality-agnostic intrinsic instance semantics. 
The enhanced modality-invariant fusion feature $F_{enh}^{in}$ is then obtained by aggregating textual values:
\begin{equation}
F_{enh}^{in} = F_{fu}^{in} + A_{fu} \odot {E_{val}}.
\end{equation}
This enhancement enforces semantically consistent invariant features across modalities, facilitating more reliable and fine-grained controllable fusion within the continuous fusion space.
To preserve consistency in non-instance regions, we further impose the background reconstruction constraint:
\begin{equation}
\begin{split}
L_{bg} & = \frac{1}{|M_b|} ( \| M_b \odot (I_{fu} - I_{ir}) \|_2^2 + \| M_b \odot (I_{fu} - I_{vi}) \|_2^2 \\
& + \mu \| M_b \odot (\nabla I_{fu} - \max(\nabla I_{ir}, \nabla I_{vi})) \|_1 ),
\end{split}
\label{eq19}
\end{equation}
where $M_b$ denotes the binary background mask. $\mu$ is the hyperparameter. The first and second terms constrain the fused image to preserve pixel content from both modalities in a balanced manner in non-instance regions, while the third term encourages the fused image to retain salient edge structures and textures from both modalities as much as possible in these regions. 

The overall objective in Training Stage \Rmnum{2} is:
\begin{equation}
    {L_{stage2}} = {L_{ROI}} + {L_{bg}}.
\end{equation}

\subsection{User Intent Querying and Fused Image Generating}
During inference, fusion is formulated as the user intent querying and fused image generating process over the learned continuous fusion space.
Given user inputs, an LLM parses both source images and user intents into structured instance–intensity (I-N) pairs and instance texts. These are further processed by Grounded-SAM and the CLIP text encoder to obtain the spatial modulation mask $M$ and text embeddings $E$, respectively.
The mask $M$ modulates modality-specific features, while $E$ enhances invariant features, enabling spatially and semantically guided fusion.
The model then generates the fused image that is well aligned with user intent.
\section{Experiments}
In this section, we evaluate ConFusion on multiple fusion datasets through qualitative and quantitative comparisons with state-of-the-art methods. Ablation studies are then performed to validate the effectiveness of the proposed modules and loss functions. Finally, we analyze the performance on object detection and further demonstrate the effectiveness of fine-grained instance-level controllable image fusion through both quantitative metrics and visualizations.

\subsection{Implementation Details and Datasets}
\noindent\textbf{Implementation Details.}
For disentangled representation learning, the network was trained for $50$ epochs using the Adam optimizer with an initial learning rate of $1\mathrm{e}^{{-4}}$ and batch size of $2$. Input images were cropped to $224 \times 224$. The loss weights were set as $\lambda = 5$, ${\alpha _1} = 0.2$ and ${\alpha _2} = 5$.
In the feature modulation and enhancement fusion, encoders and decoders were fine-tuned with Adam at a learning rate of $1\mathrm{e}^{{-6}}$, while other modules were optimized with a learning rate of $1\mathrm{e}^{{-4}}$. Training was performed for $50$ epochs with batch size of $2$. The hyperparameters were set as ${\alpha _{bg}} = 0.5$ and $\mu = 5$.
Experiments were performed using an NVIDIA RTX $4090$ GPU and an AMD EPYC $7453$ $28$-core CPU, with PyTorch $2.9.1$.

\noindent\textbf{Datasets.}
The datasets used in our experiments are derived from three widely adopted IVF benchmarks: M\textsuperscript{3}FD~\cite{tardal}, RoadScene~\cite{u2fusion}, and TNO~\cite{tno}, both providing clean and well-aligned visible-infrared image pairs. M\textsuperscript{3}FD contains 4200 pairs, split into 3900 for training and 300 for testing. RoadScene contains 220 pairs, and TNO contains 37 pairs, which are both utilized for inference.

\noindent\textbf{Metric.}
We adopt peak signal to noise ratio (PSNR)~\cite{psnr}, mean squared error (MSE)~\cite{mse}, average gradient (AG)~\cite{ag}, spatial frequency (SF)~\cite{sf}, correlation coefficient (CC)~\cite{cc} and Quality of gradient-based fusion (Q\textsuperscript{AB/F})~\cite{qabf} as metrics. Higher values of PSNR, AG, SF, CC, Q\textsuperscript{AB/F} indicate higher quality of fused images. Besides, the lower value of MSE indicate the higher quality.


\begin{table}[]
    \centering
    \caption{Quantitative comparison with state-of-the-art methods on M\textsuperscript{3}FD, RoadScene, and TNO datasets. The best and second-best results are highlighted in red and blue, respectively.}
    \label{tab1}
    \begin{adjustbox}{width=0.9\linewidth,center}
    \begin{tabular}{cc|cccccc}
        \toprule
        & Method & PSNR$\uparrow$ & MSE$\downarrow$ & AG$\uparrow$ & SF$\uparrow$ & CC$\uparrow$ & Q\textsuperscript{AB/F}$\uparrow$ \\
        \midrule
        \multirow{8}{*}{\rotatebox{90}{M\textsuperscript{3}FD}} 
        & RFNNest~\cite{rfnnest} & $\mathbf{\textcolor{blue}{63.041}}$ & $\textcolor{blue}{\mathbf{0.036}}$ & $4.287$ & $11.575$ & $\mathbf{\textcolor{blue}{0.521}}$ & $0.534$ \\
        & TarDAL~\cite{tardal} & $61.816$ & $0.048$ & $4.259$ & $12.759$ & $0.458$ & $0.409$ \\
        & CddFuse~\cite{cddfuse} & $60.838$ & $0.064$ & $\mathbf{\textcolor{blue}{5.446}}$ & $\mathbf{\textcolor{blue}{16.542}}$ & $0.430$ & $0.647$ \\
        & CoCoNet~\cite{coconet} & $62.864$ & $0.041$ & $4.417$ & $13.830$ & $0.466$ & $0.557$ \\
        & TDFusion~\cite{Lossfusion} & $62.858$ & $0.038$ & $5.382$ & $16.075$ & $\mathbf{\textcolor{blue}{0.521}}$ & $0.667$ \\
        & TextFusion~\cite{textfusion} & $61.311$ & $0.060$ & $5.181$ & $15.331$ & $0.471$ & $0.582$ \\
        & RISFuse~\cite{risfuse} & $61.767$ & $0.053$ & $5.501$ & $15.791$ & $0.392$ & $\mathbf{\textcolor{red}{0.696}}$ \\
        & Ours & $\mathbf{\textcolor{red}{63.380}}$ & $\mathbf{\textcolor{red}{0.034}}$ & $\mathbf{\textcolor{red}{5.514}}$ & $\mathbf{\textcolor{red}{17.124}}$ & $\mathbf{\textcolor{red}{0.525}}$ & $\mathbf{\textcolor{blue}{0.673}}$ \\
        \midrule
        \multirow{8}{*}{\rotatebox{90}{RoadScene}} 
        & RFNNest~\cite{rfnnest} & $62.220$ & $0.043$ & $4.082$ & $10.774$ & $0.642$ & $0.341$ \\
        & TarDAL~\cite{tardal} & $62.743$ & $0.039$ & $4.613$ & $12.153$ & $0.586$ & $0.430$ \\
        & CddFuse~\cite{cddfuse}& $61.666$ & $0.053$ & $6.001$ & $16.327$ & $0.594$ & $0.537$ \\
        & CoCoNet~\cite{coconet} & $61.623$ & $0.051$ & $5.869$ & $\mathbf{\textcolor{red}{17.104}}$ & $0.576$ & $0.371$ \\
        & TDFusion~\cite{Lossfusion} & $\mathbf{\textcolor{blue}{63.530}}$ & $\mathbf{\textcolor{blue}{0.033}}$ & $\mathbf{\textcolor{blue}{6.380}}$ & $16.631$ & $\mathbf{\textcolor{blue}{0.643}}$ & $\mathbf{\textcolor{red}{0.603}}$ \\
        & TextFusion~\cite{textfusion} & $62.969$ & $0.044$ & $3.855$ & $10.538$ & $0.619$ & $0.394$ \\
        & RISFuse~\cite{risfuse} & $62.463$ & $0.046$ & $5.401$ & $15.209$ & $0.539$ & $0.527$ \\
        & Ours & $\mathbf{\textcolor{red}{64.275}}$ & $\mathbf{\textcolor{red}{0.029}}$ & $\mathbf{\textcolor{red}{6.430}}$ & $\mathbf{\textcolor{blue}{16.732}}$ & $\mathbf{\textcolor{red}{0.646}}$ & $\mathbf{\textcolor{blue}{0.572}}$ \\
        \midrule
        \multirow{8}{*}{\rotatebox{90}{TNO}} 
        & RFNNest~\cite{rfnnest}& $63.367$ & $0.035$ & $3.969$ & $9.822$ & $0.567$ & $0.400$ \\
        & TarDAL~\cite{tardal} & $63.126$ & $0.039$ & $3.967$ & $10.857$ & $0.508$ & $0.401$ \\
        & CddFuse~\cite{cddfuse} & $61.361$ & $0.052$ & $\mathbf{\textcolor{blue}{4.985}}$ & $\mathbf{\textcolor{blue}{13.677}}$ & $0.511$ & $0.534$ \\
        & CoCoNet~\cite{coconet} & $\mathbf{\textcolor{blue}{63.791}}$ & $0.035$ & $3.003$ & $7.996$ & $0.492$ & $0.343$ \\
        & TDFusion~\cite{Lossfusion} & $63.635$ & $\mathbf{\textcolor{blue}{0.033}}$ & $4.888$ & $12.968$ & $\mathbf{\textcolor{blue}{0.569}}$ & $\mathbf{\textcolor{blue}{0.548}}$ \\
        & TextFusion~\cite{textfusion} & $62.737$ & $0.042$ & $4.307$ & $11.467$ & $0.513$ & $0.483$ \\
        & RISFuse~\cite{risfuse} & $62.527$ & $0.044$ & $4.780$ & $12.974$ & $0.463$ & $\mathbf{\textcolor{red}{0.596}}$ \\
        & Ours & $\mathbf{\textcolor{red}{64.204}}$ & $\mathbf{\textcolor{red}{0.030}}$ & $\mathbf{\textcolor{red}{5.557}}$ & $\mathbf{\textcolor{red}{14.336}}$ & $\mathbf{\textcolor{red}{0.578}}$ & $0.504$ \\
        \bottomrule
    \end{tabular}
    \end{adjustbox}
\vspace{-0.3cm}
\end{table}

\subsection{Comparison with SOTA}
The comparison methods include RFNNest~\cite{rfnnest} (INFUS, 2021), TarDAL~\cite{tardal} (CVPR, 2022), CddFuse~\cite{cddfuse} (CVPR, 2023), CoCoNet~\cite{coconet} (IJCV, 2024), TDFusion~\cite{Lossfusion} (CVPR, 2025), TextFusion~\cite{textfusion} (INFUS, 2025), and RISFuse~\cite{risfuse} (ICCV, 2025). Among these, the first five perform conventional non-controllable image fusion, while TextFusion, RISFuse, and ours belong to coarse-controllable fusion. The first five methods only use dual-modal images as input. 
For TextFusion and RISFuse, we follow their original settings with image-specific text inputs, where texts are manually crafted based on the target instances in each pair of source images.
In contrast, our method adopts a unified intent that guides the LLM to parse source images and assign instance-wise modulation intensities.


\noindent\textbf{Quantitative Evaluation.} 
To validate the competitiveness of our method, quantitative evaluations are conducted on three benchmark datasets using six widely adopted metrics, as reported in Table~\ref{tab1}. 
It can be observed that the proposed ConFusion achieves superior performance on most metrics. 
In particular, the best PSNR, MSE, and CC scores indicate that our method effectively preserves pixel-level information from source images. The AG results further demonstrate that our fused images retain richer gradient details compared to existing methods.
Moreover, the consistently excellent performance across all datasets highlights the robustness and generalization ability of our method.
Despite these favorable results, our method does not consistently achieve the best Q\textsuperscript{AB/F} and SF scores on some datasets. This can be attributed to the intent-guided inference mechanism, which emphasizes thermal information within relevant regions specified by user input. Such region-aware enhancement inevitably weakens visible texture details, leading to reduced global gradient consistency and spatial frequency. Nevertheless, ConFusion maintains competitive overall performance while offering improved controllability.

\begin{figure}
\includegraphics[width=0.9\linewidth]{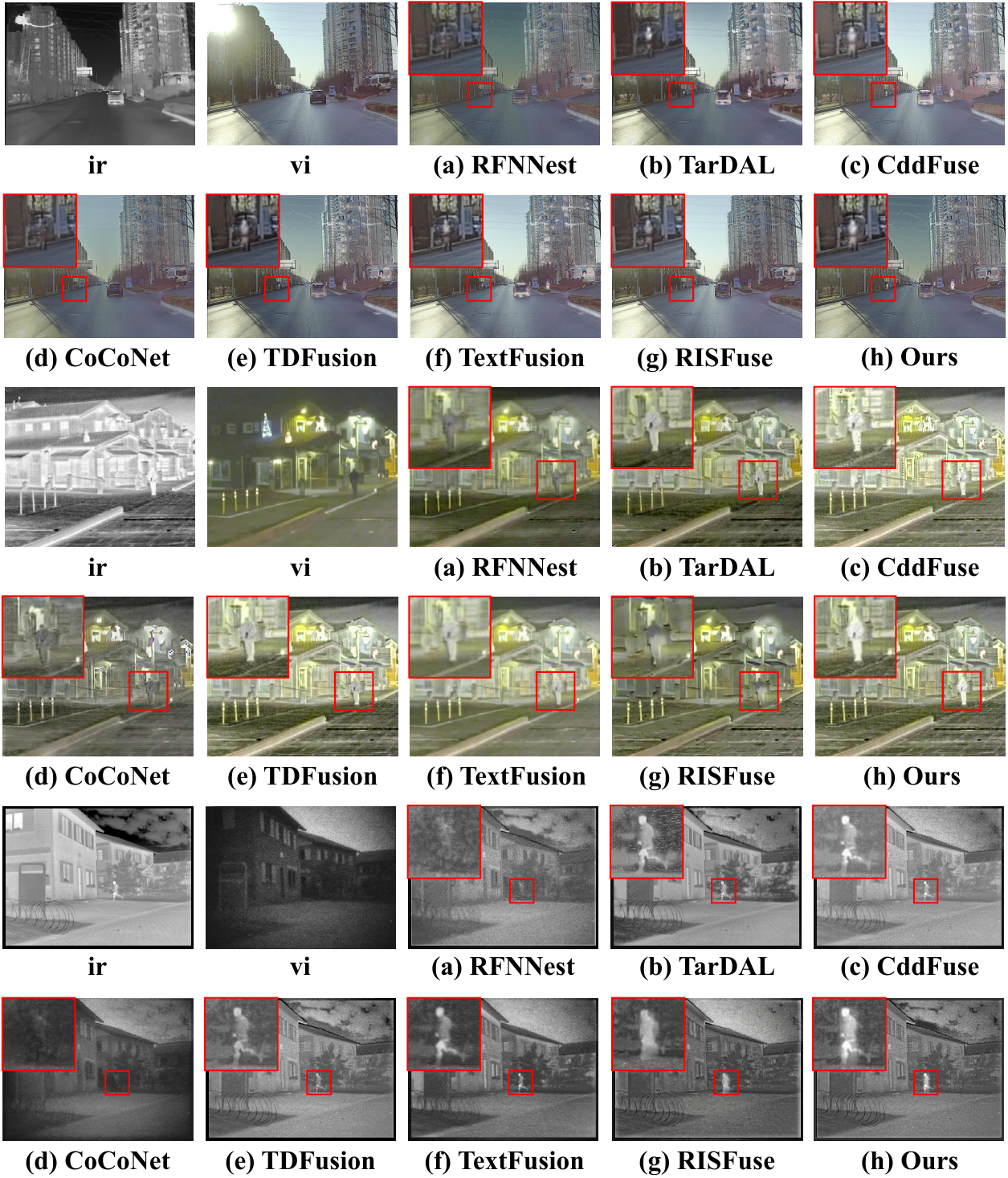}
\caption{Visualization comparison with seven state-of-the-art methods.}
\label{fig3}

\end{figure}

\noindent\textbf{Qualitative Evaluation.} Figure~\ref{fig3} illustrates qualitative comparisons on three samples from different test sets. 
Compared with existing methods, the proposed ConFusion produces fused results with clearer textures and better-preserved structural details.
For instance, structural elements such as the brick patterns on the night road, as well as the window and door edges of the house, are more distinctly preserved.
In addition, our ConFusion method produces more visually natural fusion images across most regions while preserving richer scene content. As shown in Figure~\ref{fig3}, whether it is the interior lights of the house or the color of the road, they are more consistent with the visible modality, better aligning with human perception. 
Moreover, ConFusion highlights salient thermal instances (e.g., persons) more prominently, exhibiting higher intensity responses than both non-controllable and coarse-controllable baselines. 
This is attributed to the instance-level fine-grained modulation in ConFusion, which allows more precise enhancement of semantically relevant regions through target-aware modulation intensities derived from LLM.

\begin{figure*}
\includegraphics[width=0.9\linewidth]{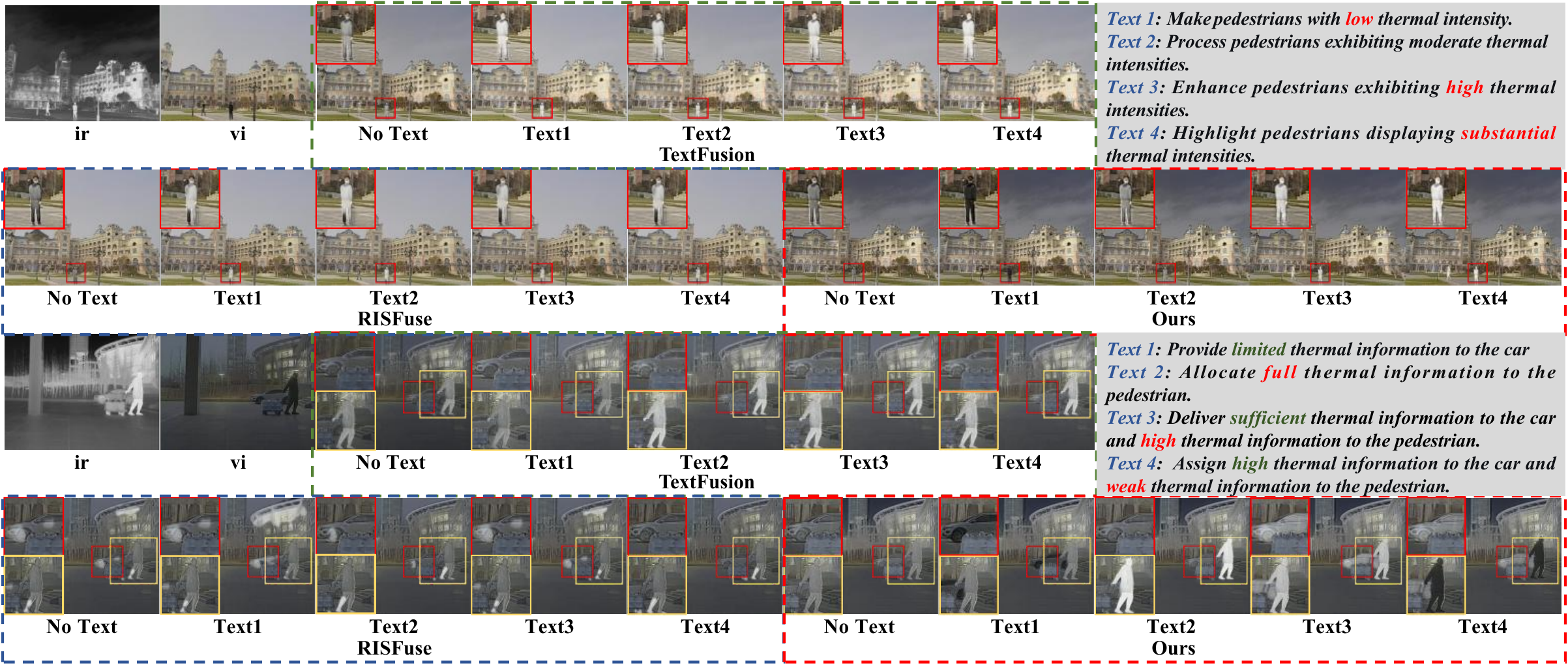}
\caption{Visualization of instance-level fine-grained controllable fusion.}
\label{fig4}
\vspace{-0.2cm}
\end{figure*}

\subsection{Ablation Studies}
As shown in Table~\ref{tab2}, we present ablation studies to assess the contributions of various components and loss functions in our ConFusion approach.

\noindent\textbf{Ablation on Invariant Semantic Alignment.}
To evaluate the role of invariant semantic alignment, we remove the $L_{text}$ during training, which aligns invariant features with CLIP-derived modality-agnostic text semantics.
As shown in Table~\ref{tab2} ($Method$ $\Rmnum{1}$ $vs.$ $Ours$), removing this constraint leads to performance degradation, highlighting its importance in disentanglement representation learning. Without invariant semantic alignment, the model fails to maintain cross-modal consistency of invariant features, ultimately impairing fusion quality.


\noindent\textbf{Ablation on Mask-Guided Specific Feature Modulator (MGSM).}
To evaluate the role of MGSM, we remove the spatial-aware modulation masks, thereby disabling instance-level guided feature modulation. The $L_{bg}$ loss treats all regions uniformly, without distinguishing between instance and non-instance areas. 
As shown in Table~\ref{tab2} ($Method$ $\Rmnum{2}$ $vs.$ $Ours$), removing MGSM slightly improves AG and SF, due to the emphasis on gradient and texture information. However, performance decreases on the remaining metrics, suggesting weaker preservation of semantically relevant regions.
In contrast, introducing MGSM leads to more balanced fusion performance and improved semantic consistency. Moreover, MGSM is essential for achieving instance-level controllability, as further validated in Sec.~\ref{secvali}.


\noindent\textbf{Ablation on Text-Driven Invariant Feature Enhance (TDIE).}
As shown in Table~\ref{tab2} ($Method$ $\Rmnum{3}$ $vs.$ $Ours$), removing the TDIE module leads to consistent degradation across all evaluation metrics.
These results highlight the importance of TDIE in improving semantic consistency and overall fusion quality.


\begin{table}[]
    \centering
    \caption{Ablation studies results about components and loss functions on M\textsuperscript{3}FD. The best results are highlighted in bold. (\Rmnum{1}): w/o \textit{Invariant Semantic Alignment}, (\Rmnum{2}): w/o \textit{MGSM}, (\Rmnum{3}): w/o \textit{TDIE},
    (\Rmnum{4}): w/o $L_{cr}$, (\Rmnum{5}): w/o $L_{dis}$, (\Rmnum{6}): w/o $L_{inv}$.}
    \begin{adjustbox}{width=0.9\linewidth,center}
    \begin{tabular}{c c c c c c c}
    \hline
       Methods & PSNR$\uparrow$ & MSE$\downarrow$ & AG$\uparrow$ & SF $\uparrow$ & CC$\uparrow$ & Q\textsuperscript{AB/F}$\uparrow$ \\
    \cline{1-7}

           \Rmnum{1} & $63.138$ & $0.036$ & $5.403$ & $16.822$ & $0.521$ & $0.665$ \\
     \Rmnum{2} & $61.678$ & $0.047$ & $\mathbf{7.539}$ & $\mathbf{22.875}$ & $0.489$ & $0.528$ \\
       \Rmnum{3} & $62.014$ & $0.044$ & $5.315$ & $17.002$ & $0.517$ & $0.606$ \\
       \cline{1-7}
              \Rmnum{4} & $61.623$ & $0.049$ & $5.146$ & $16.451$ & $0.509$ & $0.671$ \\
       \Rmnum{5} & $62.039$ & $0.044$ & $5.410$ & $17.108$ & $0.517$ & $0.653$ \\
       \Rmnum{6} & $62.072$ & $0.043$ & $5.496$& $16.983$  & $0.514$ & $0.667$ \\

    \cline{1-7}
       Ours & $\mathbf{63.380}$ & $\mathbf{0.034}$ & $5.514$ & $17.124$ & $\mathbf{0.525}$ & $\mathbf{0.673}$\\
    \hline
    \end{tabular}
    \end{adjustbox}
    \vspace{-0.2cm}
    \label{tab2}

\end{table}

\noindent\textbf{Ablation on Loss Functions.}
We further evaluate the roles of $L_{cr}$, $L_{dis}$, and $L_{inv}$ by removing each term. 
As shown in Table~\ref{tab2} ($Method$ $\Rmnum{4}$, $\Rmnum{5}$, and  $\Rmnum{6}$ $vs.$ $Ours$), removing any of these losses consistently degrades performance, indicating necessity for effective feature disentanglement. 
Specifically, $L_{cr}$ enforces cross-reconstruction to preserve complementary information, $L_{dis}$ promotes the decoupling between modality-invariant and modality-specific features, and $L_{inv}$ maintains semantic consistency of invariant features. 
Together, these losses enable more effective disentanglement representation learning, leading to improved fusion performance.

\begin{figure*}
\includegraphics[width=0.9\linewidth]{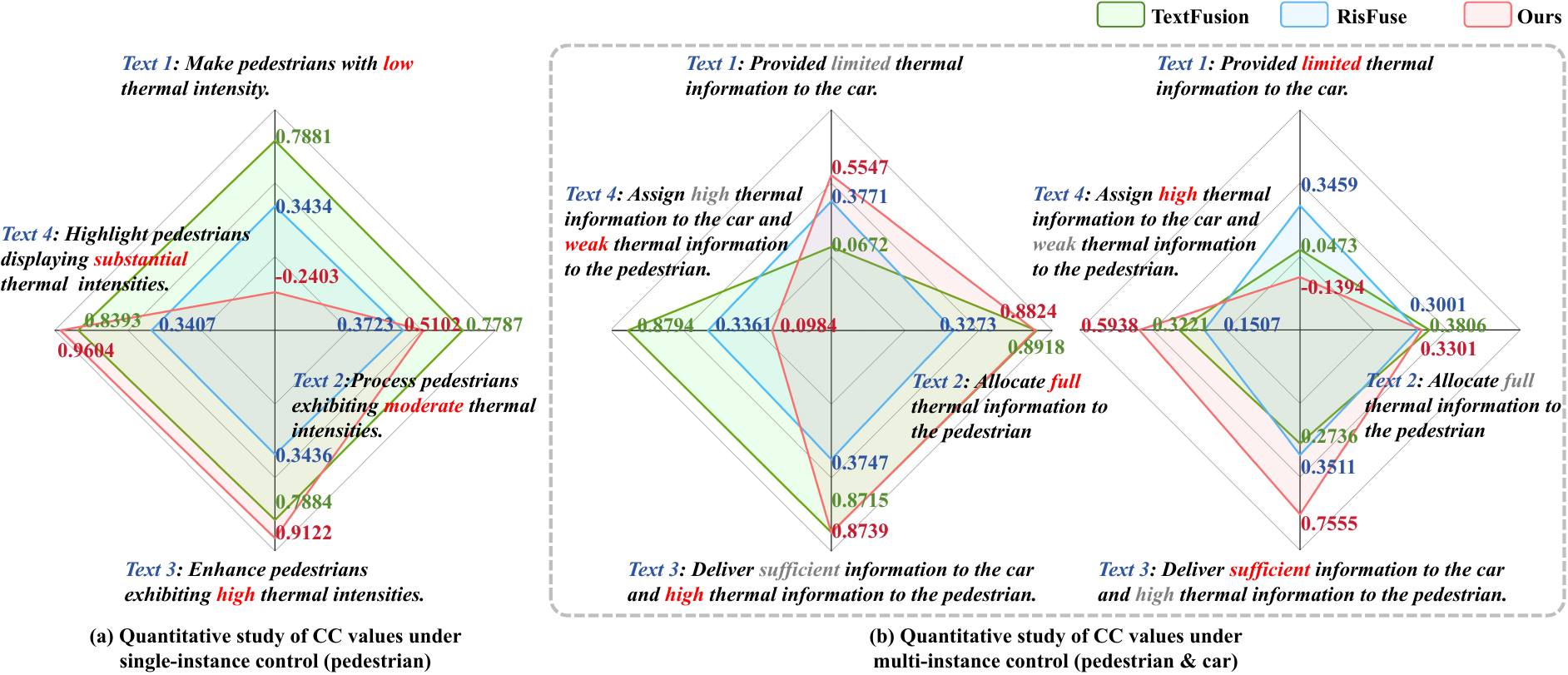}
\caption{Quantitative study of Instance-Level Fine-Grained Controllable Fusion, where CC denotes correlation coefficient between fused and infrared images in instance regions under diverse user intents.}
\label{fig5}
\end{figure*}

\begin{table*}[]
    \centering
    \caption{Quantitative comparison of object detection performance on M\textsuperscript{3}FD. The best and second-best results are highlighted in red and blue, respectively.}
    \begin{adjustbox}{width=0.72\linewidth,center}
    \begin{tabular}{c c c c c c c | c c c c}
    \hline
       \multirow{2}* {Methods} & \multicolumn{6}{c|}{Ap50} & \multirow{2} * {precision} & \multirow{2} * {recall} & \multirow{2} * {mAP50} & \multirow{2} * {mAP50:95} \\
    \cline{2-7}
       ~ & car & person & bus & truck & motor & lamp & ~ &  & ~ & ~ \\
    \cline{1-11}
        RFNNest~\cite{rfnnest} & $0.927$ & $0.833$ & $0.903$ & $\mathbf{\textcolor{blue}{0.872}}$ & $0.770$ & $0.819$ & $0.898$ & $0.784$ & $\mathbf{\textcolor{blue}{0.854}}$ & $0.583$ \\
        TarDAL~\cite{tardal} & 0.923  & $\mathbf{\textcolor{red}{0.858}}$& $\mathbf{\textcolor{red}{0.912}}$ & $0.862$ & $0.770$ & $0.779$ & $0.891$ & $0.783$ & $0.851$ & $0.582$ \\
        CddFuse~\cite{cddfuse} & $0.929$ & $\mathbf{\textcolor{blue}{0.853}}$ & $0.886$ & $0.844$ & $0.747$ & $0.813$ & $0.890$ & $0.779$ & $0.845$ & $0.583$ \\
        CoCoNet~\cite{coconet} & $0.928$ & $0.820$ & $0.887$ & $0.857$ & $0.753$ & $0.819$ & $0.899$ & $0.786$ & $0.844$ & $0.580$ \\
        TDFusion~\cite{Lossfusion} & $\mathbf{\textcolor{blue}{0.930}}$ & $0.852$ & $0.880$ & $0.865$ & $\mathbf{\textcolor{blue}{0.776}}$ & $\mathbf{\textcolor{blue}{0.820}}$ & $\mathbf{\textcolor{blue}{0.905}}$ & $\mathbf{\textcolor{blue}{0.792}}$ & $\mathbf{\textcolor{blue}{0.854}}$ & $\mathbf{\textcolor{blue}{0.592}}$ \\
        TextFusion~\cite{textfusion} & $0.927$ & $0.842$ & $0.897$ & $0.865$ & $0.768$ & $0.806$ & $0.888$ & $0.783$ & $0.851$ & $0.584$\\
        RISFuse~\cite{risfuse} & $\mathbf{\textcolor{red}{0.933}}$ & $0.834$ & $\mathbf{\textcolor{blue}{0.905}}$ & $0.845$ & $0.775$ & $\mathbf{\textcolor{red}{0.828}}$ & $0.900$ & $0.788$ & $0.853$ & $0.583$\\
        Ours & $\mathbf{\textcolor{blue}{0.932}}$ & $0.844$ & $0.904$ & $\textbf{\textcolor{red}{0.882}}$ & $\mathbf{\textcolor{blue}{0.777}}$ & $\mathbf{\textcolor{blue}{0.826}}$ & $\mathbf{\textcolor{red}{0.906}}$ & $\mathbf{\textcolor{red}{0.794}}$ & $\mathbf{\textcolor{red}{0.861}}$ & $\textbf{\textcolor{red}{0.593}}$ \\
    \hline
    \end{tabular}
    \end{adjustbox}
    \label{tab3}
\end{table*}

\subsection{Analysis of Instance-Level Fine-Grained Controllable Fusion}
\label{secvali}
We evaluate the fine-grained controllability of ConFusion under diverse user intents.
As illustrated in the first two rows of Figure~\ref{fig4}, we modulate the thermal intensity within the pedestrian region according to text inputs.  
Although TextFusion and RISFuse can adjust the thermal responses, their behaviors remain coarse and fail to reflect subtle differences in intents. 
For instance, both methods increase thermal intensity even for “low” and produce nearly indistinguishable results for “high” and “substantial”.
In contrast, ConFusion generates clearly distinguishable thermal responses that align well with the intended semantics, demonstrating fine-grained controllability at the instance-level.
For quantitative evaluation, Figure \ref{fig5}(a) reports the CC values between fused and infrared images within the pedestrian region, corresponding to Figure~\ref{fig4}. 
The CC values of TextFusion and RISFuse remain nearly unchanged across inputs such as “low”, “moderate”, “high”, and “substantial”, indicating limited controllability.
In comparison, ConFusion shows a clear and consistent correspondence with the input intents.
Specifically, “substantial” and “high” correspond to the highest and second-highest CC values, respectively, further demonstrating its fine-grained control capability.

Furthermore, the proposed ConFusion effectively handles multi-instance scenarios.
As shown in the last two rows of Figure~\ref{fig4}, when user intents specify different requirements for multiple instances, TextFusion and RISFuse fail to independently adjust each instance.
For example, when Text $2$ instructs the pedestrian to exhibit full thermal intensity, TextFusion enhances both car and pedestrian instances. When Text $3$ instructs the pedestrian and car to exhibit sufficient and high thermal intensity respectively, RISFuse fails to produce the desired responses for either instance. 
In contrast, the proposed ConFusion produces results consistent with the specified intents, allowing different instances to be adjusted independently or jointly. For instance, when Text $4$ instructs high thermal intensity for the pedestrian but low for the car, our method successfully generates the corresponding fusion result. 
Correspondingly, Figure~\ref{fig5}(b) shows that the CC values of our method closely match the specified intents for both instances, while remaining at a moderate level when no explicit instruction is given. These results further demonstrate that our method effectively achieves instance-level fine-grained controllable fusion.

\subsection{Performance on Object Detection}
Table~\ref{tab3} reports object detection results on the fused images from M\textsuperscript{3}FD dataset. 
We resplit the dataset by moving $750$ images from the training set to the test set, resulting in the $3:1$ train-test split.
We adopt YOLOv$9$~\cite{yolov9} as the detector and train for $400$ epochs.
ConFusion achieves competitive performance across categories, obtaining the best result on the truck class and the second-best on car, motor, and lamp in mAP${50}$.
Furthermore, it achieves the best overall performance across all categories in mAP${50}$, mAP$50$:$95$, Precision, and Recall.
These results indicate that the proposed instance-level fine-grained controllability improves semantic representation quality, thereby benefiting downstream detection performance.


\subsection{Conclusion}
In this paper, we propose the ConFusion framework that learns a continuous fusion space via Gaussian-conditioned spatial-aware modulation, enabling instance-level fine-grained controllable infrared–visible image fusion driven by user intent.
A dual-branch architecture is first employed to disentangle modality-invariant and modality-specific representations, guided by joint reconstruction and text-guided semantic alignment. Fine-grained instance-level modulation is achieved through the Mask-Guided Specific Feature Modulator, which leverages Gaussian-conditioned instance variables together with instance masks from Grounded SAM, while the Text-Driven Invariant Feature Enhancer improves semantic consistency and fusion quality.
By mapping continuously sampled instance-level variables to spatial modulation, the model forms a continuous fusion space, which supports flexible and fine-grained controllable Infrared and Visible Image Fusion.
During inference, user intents are parsed by an LLM into instance-level modulation variables to steer the fusion process. Extensive experiments demonstrate that ConFusion achieves superior fusion quality and downstream task performance compared to state-of-the-art methods, while supporting fine-grained controllable image fusion.

Despite the promising results, ConFusion has several limitations. First, the incorporation of an LLM and Grounded SAM introduces additional computational overhead. 
Second, the performance depends on accurate instance segmentation and reliable interpretation of user intent by the LLM. In the future, more efficient and lightweight frameworks can be explored, along with improved robustness to segmentation errors and ambiguous user intent.

\bibliographystyle{ACM-Reference-Format}
\bibliography{sample-base}

\appendix
\end{sloppypar}
\end{document}